%% file: main.tex
\documentclass[letterpaper, 10 pt, conference]{ieeeconf}  

\IEEEoverridecommandlockouts                              

\usepackage{graphics} 
\usepackage{epsfig} 
\usepackage{caption}
\usepackage{subfigure}
\usepackage{bm}
\usepackage{amsmath} 
\usepackage{amssymb}  
\usepackage{lipsum}
\usepackage{comment}
\usepackage{xcolor}
\usepackage{algorithm,algorithmic}
\usepackage{relsize}
\usepackage{xurl}
\usepackage{booktabs}
\usepackage{multirow}
\usepackage{siunitx}
\makeatletter
\let\NAT@parse\undefined
\makeatother
\usepackage{hyperref}

\newtheorem{theorem}{Theorem}[section]
\newcommand{\projectName}{GaPT}

\title{\LARGE \bf
GaPT: Gaussian Process Toolkit for Online Regression with Application to Learning Quadrotor Dynamics
}

\author{Francesco Crocetti$^{1,2*}$, Jeffrey Mao$^{1*}$, Alessandro Saviolo$^{1}$, Gabriele Costante$^{2}$, and Giuseppe Loianno$^{1}$
\thanks{$^*$These authors contributed equally and listed alphabetic order.}
\thanks{$^{1}$The authors are with the New York University, Tandon School of Engineering, Brooklyn, NY 11201, USA.
        {\tt\footnotesize email:\{fc2353, jm7752, as16054, loiannog\}@nyu.edu.}}
\thanks{$^{2}$The authors are with the Department of Engineering, University of Perugia, 06125 Perugia, Italy. 
{\tt\footnotesize email:\{francesco.crocetti, gabriele.costante\}@unipg.it.}}
\thanks{This work was supported in part by the NSF CAREER Award 2145277, the DARPA YFA Grant D22AP00156-00, the Technology Innovation Institute, Qualcomm Research, Nokia, and NYU Wireless. Giuseppe Loianno
serves as consultant for the Technology Innovation Institute. This arrangement
has been reviewed and approved by the New York University in accordance
with its policy on objectivity in research.}
}

\begin{document}

\maketitle
\thispagestyle{empty}
\pagestyle{empty}

\begin{abstract}
Gaussian Processes (GPs) are expressive models for capturing signal statistics and expressing prediction uncertainty. As a result, the robotics community has gathered interest in leveraging these methods for inference, planning, and control. 
Unfortunately, despite providing a closed-form inference solution, GPs are non-parametric models that typically scale cubically with the dataset size, hence making them difficult to be used especially on onboard Size, Weight, and Power (SWaP) constrained aerial robots. In addition, the integration of popular libraries with GPs for different kernels is not trivial. 
In this paper, we propose GaPT, a novel toolkit that converts GPs to their state space form and performs regression in linear time.
GaPT is designed to be highly compatible with several optimizers popular in robotics.
We thoroughly validate the proposed approach for learning quadrotor dynamics on both single and multiple input GP settings.
GaPT accurately captures the system behavior in multiple flight regimes and operating conditions, including those producing highly nonlinear effects such as aerodynamic forces and rotor interactions.
Moreover, the results demonstrate the superior computational performance of GaPT compared to a classical GP inference approach on both single and multi-input settings especially when considering large number of data points, enabling real-time regression speed on embedded platforms used on SWaP-constrained aerial robots.
\end{abstract}

\section*{Supplementary Material}
\noindent \textbf{Code}: \url{https://github.com/arplaboratory/GaPT}\newline
\noindent \textbf{Video}: \url{https://youtu.be/Bi5x2sQcW3s}


\input{Sections/I_introduction}

\input{Sections/II_related_works}

\input{Sections/III_gaussian_process}

\input{Sections/IV_metodology}
\input{Sections/timing_table_average}

\input{Sections/V_experimental_setup}

\input{Sections/pred_performance_fig}
\input{Sections/VI_results}

\input{Sections/VII_conclusions}

\clearpage

\bibliographystyle{IEEEtran}
\bibliography{biblio}







\end{document}

%% file: Sections/I_introduction.tex
\section{Introduction}\label{sec_I}
Gaussian Processes (GPs) have become popular models in a wide range of robotics inference~\cite{Guillem_datadriveMPC,hewing2019cautious,Deisenroth_GP_robotics}, planning~\cite{barfoot2014batch}, and control problems~\cite{Anayo_reachabilityonlineGP,WangonlineGP2018,loquercio2022}. 
However, their inference time scales cubically with respect to the dataset size, thus making these models unsuitable for many real-time robotics problems, especially on Size, Weight, and Power (SWaP) constrained aerial robots. 
This motivates the need to develop novel approximation techniques that can perform inference at a much higher rate without sacrificing the accuracy of the original model.
In this paper, we present \projectName, a GP Toolkit that converts a GP into a linear state space form. This form processes information sequentially to learn the data trend allowing the computation to scale linearly with the dataset instead. 
\projectName~can easily instantiate any kernel or combination of them in C++ with minor or no user effort. 
The approach does not perform the expensive matrix inversions or multiplications of a typical GP and is easily portable to  SWaP-constrained robots.



 \begin{figure}
     \centering
     \vspace{0.7em}
     \includegraphics[width=\linewidth, trim=0 0 0 0, clip]{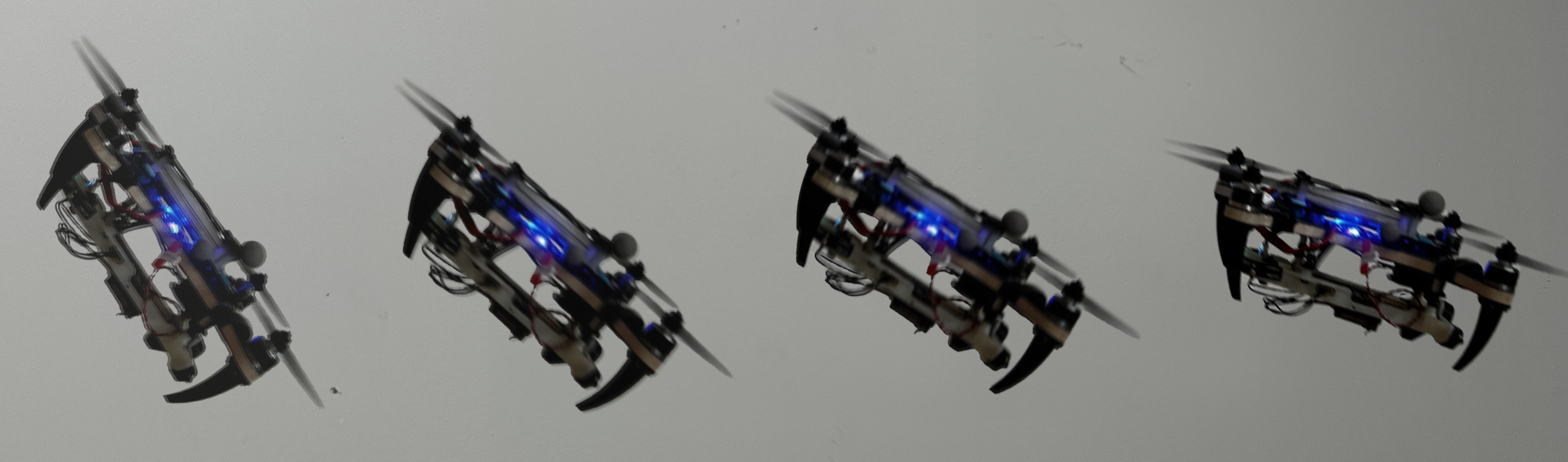}\\
     \includegraphics[width=1.3\linewidth, trim=350 500 0 0, clip]{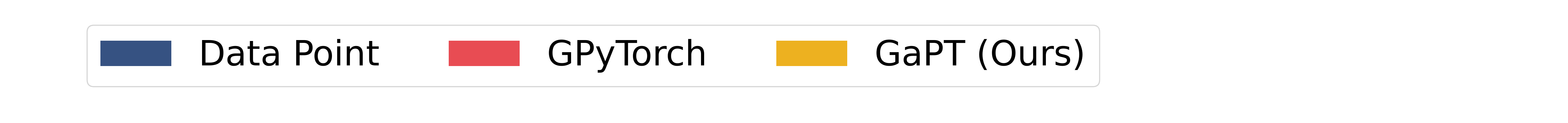}\\
     \includegraphics[width=\linewidth, trim=0 0 0 0, clip]{images/First_Page/siso_KF_pred_test_1.png}
     \caption{
     Comparison between GaPT and a classical GP inference approach (GPyTorch) for learning quadrotor residual dynamics during agile flight on $6000$ test data points.
     }
     \label{fig:intro_fig}
     \vspace{-2em}
\end{figure}

Our main contributions are the following. First, we design and present \projectName, a GP Toolkit that can perform inference at a much faster rate than the standard inference model used in popular libraries such as~\cite{gardner2018gpytorch} by translating GPs into a linear state space form. 
The linear state space form or State Space Gaussian Process (SSGP) scales linearly with respect to the dataset size. This allows quicker data processing and overcomes one of the major hurdles related to real-time robotics applications, especially when employing large datasets on SWaP constrained aerial robots. In fact, the histogram in Fig.~\ref{fig:intro_fig} shows the superior computational advantages of GaPT which requires $\times 37$ less time compared to the GP baseline, while also producing similar inference results when estimating external disturbances.
Second, \projectName~ includes the state space realization for Single Input Single Output (SISO) as well as for Multiple Input Single Output (MISO) robotics systems. 
By expanding the input space to multi-dimensions, \projectName~ better captures the high-frequency characteristic of the true signal, outperforming single input GPs on unfiltered real-world data, as experimentally validated in our results. 
Third, we provide a variety of ready-to-use kernel implementations such as RBF, Periodic, and Matern kernels. Fourth, we validate our approach on real-world learning quadrotor dynamics inference problems
to capture the system behavior in multiple flight regimes and operating conditions, including those producing highly nonlinear effects such as aerodynamic forces and torques, and rotor interactions. Finally, our method includes a quick training routine and can easily export C++ ready functions that can be integrated into several optimization libraries (e.g., \cite{acados,FORCESPro,mastalli20crocoddyl}) often used on small-scale aerial~\cite{Guillem_datadriveMPC} and ground~\cite{hewing2019cautious} robots. Our approach and its open source implementation will contribute to reducing the entry barrier to the adoption of these approaches avoiding the complex integration and computation time required by several libraries, especially for SWaP robots.

%% file: Sections/II_related_works.tex
\section{Related Works}

In literature, GPs have been applied to learning residual or full system dynamics~\cite{Guillem_datadriveMPC, hewing2019cautious, Deisenroth_GP_robotics}, reward functions~\cite{biyik2020active}, planning \cite{barfoot2014batch}, adaptive control \cite{gahlawat2020l1}, or learning safety regions for repeated tasks such as in reinforcement learning \cite{Anayo_reachabilityonlineGP, WangonlineGP2018}. 
These methods are often implemented by hard coding the full GP model in C++ such as in \cite{Guillem_datadriveMPC, hewing2019cautious}. 
In the realm of Bayesian Machine learning, Python libraries, such as GPytorch \cite{gardner2018gpytorch} or GPyflow \cite{matthews2017gpflow}, are useful for training hyperparameters and inference. 
These libraries leverage PyTorch or TensorFlow systems respectively for back propagation and GPU integration allowing the easy implementation of new kernels or more complex GP models \cite{SMkernerl, zhao2021deep}, but are complicated to interface with robotics optimization solvers \cite{acados,FORCESPro,mastalli20crocoddyl}. 

Being non-parametric, the inference computation time of GPs generally scales with $O(n^3)$ with $n$ the dataset size even when using the aforementioned libraries although the corresponding processing time is internally mitigated by leveraging GPU parallelism. The process of accelerating GPs is a major field of study. 
Data dimensionality reduction techniques~\cite{Sivaram_fastGP_calc, joaquin_unifiyng_view, DAS201812} attempt to reduce the dataset size to a fraction of the points while keeping similar predictive performance. However, by performing data selection, a GP can never retain its full expressiveness especially if a large amount of data has to be discarded especially if converting inference time from a cubic order to a linear order a correspondingly large amount of data must be discarded.
Our current method sidesteps this issue by converting the GP to an approximate state space form which only requires $O(n)$ time complexity allowing us to perform rapid inference without discarding data. This approach is inspired by recent signal processing results~\cite{sarkka2011linear,  corenflos2022temporal}.
However, \cite{sarkka2011linear} derives the state space form but only for a SISO system and is tailored for signal processing problems. In addition, it is very difficult to obtain a step-by-step formulation and portable solution tailored for small-scale robots.
For further computation time increases among state space models, \cite{corenflos2022temporal} also proposes a framework for SSGPs, but the code relies heavily on pre-computing the Kalman filter predictions and updates steps to achieve these computation time gains making it less suitable for online learning.  
The aforementioned SSGP works are also not formulated for processing multiple inputs in the GPs and only consider single outputs with the exception of~\cite{sarkka2013spatiotemporal}. However, this approach is difficult to derive, implement, and practically use for dimensions beyond two. \projectName~ is easy to derive, and implement and has been experimentally validated. Finally, our framework seeks to promote the use of additional kernels by implementing several variants and provides a clear approach to enable any user to derive with minor effort additional ones based on the research needs. 


%% file: Sections/III_gaussian_process.tex
\section{Gaussian Process Regression}\label{sec_GP}

Considering a multi dimensional input $\mathbf{x} \in \mathbb{R}^n$, a Gaussian Process (GP) is stochastic process ${f}(\mathbf{x})$ where any collection  of random variables has a joint Gaussian Distribution $[f(\mathbf{x}_1),f(\mathbf{x}_2),\ldots,f(\mathbf{x}_n)] \sim \mathcal{N}(\boldsymbol {\mu}, \mathbf{K})$.
A GP is commonly defined as introducing the kernel formalism \cite{RasmussenW06}
\begin{equation}
    f(\mathbf{x}) \sim \mathit{GP}(\mu(\mathbf{x}), k(\mathbf{x}, \mathbf{x}')),
    \label{eq_GP_def}
\end{equation}
where $\mu(\mathbf{x}): \mathbb{R}^n \rightarrow \mathbb{R}$ and $k(\mathbf{x}, \mathbf{x}'): \mathbb{R}^n \times \mathbb{R}^n \rightarrow \mathbb{R}$ represent the mean and the covariance functions respectively. 
For a general regression problem, given a set of training data $\mathcal{D}=\{\mathbf{X}, \mathbf{y}\}$, with $\mathbf{X}=\{\mathbf{x}_{i} \in  \mathbb{R}^d\}_{i=1}^n$  and $\mathbf{y}=\{y_{i} \in  \mathbb{R}\}_{i=1}^n$, the interest is finding a function $\mathbf{f} : \mathbb{R}^n \rightarrow \mathbb{R}$ so that given an unknown input $\mathbf{x}_{\ast} \in \mathbb{R}^d$ the predicted value $y_\ast \in \mathbb{R}$ is reliable. In GP regression, $f(\mathbf{x})$ are assumed sampled from a zero mean Gaussian process (see eq.~(\ref{eq_GP_def})) and the observations are assumed corrupted by additive Gaussian noise \cite{sarkka2013spatiotemporal, murray2010slice}
\begin{align}
        \text{GP Prior} \quad& f(\mathbf{x_{\ast}}) \sim \mathit{GP}(0, k(\mathbf{x}, \mathbf{x}'; \boldsymbol\theta))
    \label{eq_GP_prior}\\
        \text{Measurements} \quad & y_i = f(\mathbf{x}_i) + \epsilon, \quad \epsilon \sim \mathcal{N}(0,\sigma_\text{noise}^2)
    \label{eq_GP_measurements}
\end{align}
where $\boldsymbol{\theta}$ represents the kernels hyper-parameters (length-scales, magnitude, period length, etc) which could be inferred from the marginal likelihood $\mathcal{N}(\mathbf{y}|\mathbf{0}, \mathbf{K}_{xx})$ maximizing the log-marginal likelihood
\begin{equation}
    p(\mathbf{y} | \mathbf{X}, \boldsymbol{\theta}) = -\frac{1}{2}\text{log} |\mathbf{K}|  -\frac{1}{2}\mathbf{y}^\top \mathbf{K}^{-1} - \frac{n}{2}\text{log}2\pi,
    \label{eq_log_marg_likelihood}
\end{equation}
where $\mathbf{K} = k(\mathbf{X},\mathbf{X}, \boldsymbol\theta) + \sigma_\text{noise}^2\mathbf{I}$ is the covariance matrix for the targets $\mathbf{y}$. The main advantage of employing GPs as regression models is the analytically tractable and close form solution of the predictive distribution~\cite{RasmussenW06}  which is a Gaussian distribution with mean and variance
%
%
%
\begin{equation}
    \begin{aligned}
    \mathbf{k}_\ast&=[k(\mathbf{x}_\ast,\mathbf{x}_0)~ k(\mathbf{x}_\ast,\mathbf{x}_1)\cdots ~k(\mathbf{x}_\ast,\mathbf{x}_n)]^\top,\\
   \mu_i(f(\mathbf{x}_{\ast})) &= \mathbf{k}_\ast^\top(\mathbf{K}+\sigma_{\text{noise}}^2\mathbf{I})^{-1}\mathbf{y},\\
   \mathbb{V}[f(\mathbf{x}_{\ast})] &= k(\mathbf{x_\ast},\mathbf{x_\ast})-\mathbf{k}_\ast^\top(\mathbf{K}+\sigma_{\text{noise}}^2\mathbf{I})^{-1}\mathbf{k}_\ast.
    \end{aligned}
    \label{eq_GP_prediction}
\end{equation}

Consequently, one of the main drawbacks of using Gaussian processes is the computational complexity, which can be limiting in applications where a large dataset is required to learn a complex function. The mean and variance evaluations (see Eq.~(\ref{eq_GP_prediction})) are ruled by matrix inversions. 
Even given a pre-computed inverse it still takes $O(n^3)$ due to the multiplication which limits the amount of data usable in time-sensitive tasks such as robot control.

%% file: Sections/IV_metodology.tex
\section{Methodology}
We describe a hierarchical approach to converting stationary kernel into its state space form for MISO systems and present a clear derivation of the state space form for different kernels useful in robotics.
\subsection{Dual GP representation for stationary kernels}\label{subsec_Gp_b}
Performing a proper statistical inference often requires a large number of samples. This clashes with the high computational complexity required by GP-based inference. However, for stationary kernels, the Gaussian Process can be thought of as a solution to an approximate $m{\text{-th}}$ order Stochastic Differential Equation (SDE)~\cite{hartikainen2010kalman}. 
The Linear State Space Model (LSSM) representation of the system allows us to consider the statistical inference problem as a state estimation task that can be solved with an equivalent Kalman filter. The computational cost for a single evaluation depends on the m-order of the system $O(m^3)$, and if the number of the observations $n \gg m$, the influence of $m^3$ can be neglected obtaining $O(nm^3) \sim O(n)$. $m$ in this case is a user-controlled hyperparameter  maps to a Taylor Series approximation order of the Kernel. The approach represents an advantageous alternative to the classical GP inference method (with $O(n^3)$ cost).
     Therefore, this framework takes into account some covariance functions of a stationary process (Radial Basis kernel, Matern Kernel, Periodic Kernel), where the covariance functions are input invariant to translations (i.e., $k(\mathbf{x},\mathbf{x}') = k(\Delta)$ where $\Delta = \mathbf{x}-\mathbf{x}'$). 
     All stationary kernels can be directly derived by their spectral density \cite{bochner1959lectures,stein1999interpolation}.
Because all stationary kernel functions are positive definite and stationary by the GP definition, we can apply Bochner's theorem as below to form a Fourier dual with the covariance function and its power spectrum.   %
\noindent\begin{theorem}[Bochner]
\textit{A continuous stationary function is positive definite
if and only if $k$ is the Fourier transform of a finite positive measure}
\begin{equation}
    \begin{aligned}
        k(\Delta) &=   \int \mathbf{S}(\omega)e^{2 \pi i\omega \mathbf{\Delta}} d\omega,
    \end{aligned}
\end{equation}
\textit{where $\mathbf{S}$ is a positive finite measure}.
\end{theorem}
If  $k(\Delta)$ has a density $\mathbf{S}(\omega)$ then $\mathbf{S}$ is known as the spectral density (or power spectrum) corresponding to $k$. 
In this case, the covariance function and spectral density are Fourier duals \cite{chatfield2003analysis}
\begin{equation}\label{eqn:bochner}
        \mathbf{S}(\omega) =  \int_{}^{} k(\Delta)e^{-2 \pi i\omega \mathbf{\Delta}} d\mathbf{\Delta}.
\end{equation}
\begin{algorithm}[t]
\caption{State Space Gaussian Process Conversion\label{alg_ssgp} }
\begin{algorithmic}[1]
  \STATE $\mathcal{F}[k(\tau)]=\mathbf{S}(\omega)=F(j\omega)F(-j\omega) \label{eq:fourier_kernel}$ \COMMENT{Bochner}
  \STATE $\frac{q_c^2}{\mathbf{S}(w)}=s(j\omega)s(-j\omega) \approx \sum_{i=0}^{m} c_{i}(\omega^2)^{i}\label{eq:Taylor_exp}$\COMMENT{Taylor Series}
  \STATE $\sum_{i=0}^{m} c_{i}(\omega^2)^{i}=  \prod_{i=1}^{m} (r_i^2+\omega^2)$ \COMMENT{Root Factor}
  \STATE $\sum_{i=0}^{m}c_{i}(\omega^2)^{i}=  \prod_{i=1}^{m} (r_i-j\omega)(r_i+j\omega)\label{eq:spectral_decom}$ \COMMENT{Decomp.}
  \STATE $ s(j\omega)=  \prod_{i=1}^{m} (r_i+j\omega) \label{eq:conj_comp}$ \COMMENT{Positive conjugate,}
 \STATE $s(j\omega)=  \sum_{i=0}^{m} a_{i}(j\omega)^{i} \label{eq:defactor}$ \COMMENT{Defactorization,}
 \STATE $ \mathcal{F}^{-1}\left[ s(j\omega)\right]  =  \sum_{i=0}^{m} a_{i}\frac{\partial ^{i}f(x)}{\partial  x^{i}}\label{eq:inverse_Fourier}$ \COMMENT{Inverse Fourier}
\end{algorithmic}
\end{algorithm}

\subsection{State Space Gaussian Process (SSGP)}
A Gaussian Process can be defined by two major components described in eqs.~(\ref{eq_GP_prior}) and ~(\ref{eq_GP_measurements}) Unfortunately, the GP Prior equation, eq.~(\ref{eq_GP_prior}) is in a large matrix form making it computationally complex. We turn our attention to an alternative formulation of GP Prior described solely as a differential equation driven by a white noise process (zero-mean Gaussian process with a Dirichlet kernel function of magnitude $q_c$)  \cite{sarkka2013spatiotemporal}. We start by considering a single input single output formulation in terms of a single input variable $x \in \mathbb{R} $ as
\begin{equation}  
Q(x)=  a_l \frac{\partial^l f(x)}{\partial x^l}+\cdots+a_1 \frac{\partial f(x)}{\partial x}+a_0 f(x),
\label{eqn:diff_eq}
\end{equation}
where $Q(x)$ represents a white noise process and $f(x)$ represents the function we would like to fit over our data. Conceptually, we can see the reformulation of the prior from matrix form to a differential equation as the difference in taking the averaged sum of correlations from all points in the dataset eq.~(\ref{eq_GP_prediction}) as opposed to the trend of motion described by the previous points in the dataset in eq.~(\ref{eqn:diff_eq}). 
The additional advantage of seeing the GP in terms of the trend of the previous input points as opposed to the sum of all correlations to points is the ability to process information sequentially.
We can derive the transfer function of eq.~(\ref{eqn:diff_eq}) as
\begin{equation}
    F(j\omega) = \frac{q_c}{\underbrace{a_n (j\omega)^l+\cdots+a_1 (j\omega)+a_0}_{s(j\omega)}},
    \label{eq:transfer}
\end{equation}
where $q_c \in \mathbb{R}$ is a constant describing the Fourier transform of $Q(x)$ of the process. The $q_c$ parameter can be obtained through either optimization in hyperparameter training or tuned by a user.
An important facet to emphasize is that a stationary kernel function $k(\Delta)$ is described completely and solely by its spectral density and eq.~(\ref{eq:transfer}) also has a spectral density. 
We derive the steps needed to convert an arbitrary stationary kernel function $k(\Delta)$ into a corresponding differential equation. The key component is to obtain a differential equation in the form of eq.~(\ref{eqn:diff_eq}) from $s(j\omega)$ components. 
\renewcommand{\algorithmiccomment}[1]{\bgroup\hfill~#1\egroup}

The full method to construct this relationship is described in Algorithm \ref{alg_ssgp}. First, in step (\ref{eq:fourier_kernel}), the left and right side equivalences are due to the Bochner theorem and the definition of spectral density respectively. 
Very few kernels can be represented perfectly as a linear differential equation. This necessitates an approximation with a high enough $m$ order Taylor series polynomial shown in step~(\ref{eq:Taylor_exp}) of our algorithm. 
The Taylor series expansion is in terms of $\omega^2$ to allow easy spectral decomposition in positive and negative conjugate components as in step~(\ref{eq:spectral_decom}).  Only the positive component from step~(\ref{eq:conj_comp}) is required. 
From step~(\ref{eq:conj_comp}), we can de-factorize the positive conjugate in step (\ref{eq:defactor}) to obtain the coefficients of the polynomial.  Finally, step~(\ref{eq:inverse_Fourier}) clearly shows that we obtain the form of eq.~(\ref{eqn:diff_eq}) after performing the inverse Fourier transform. 
Overall, in this process, we matched the spectral density of eq.~(\ref{eq:transfer}) with the spectral density described by eq.~(\ref{eqn:bochner}). If the two stationary GP match spectral densities, then they are the same GP. Therefore, given the linear differential equation in eq.~(\ref{eqn:diff_eq}) and $ \mathbf{f}(x)=\begin{bmatrix}
f(x) & \frac{\partial f(x)}{\partial x} &
\hdots &
\frac{\partial^m f(x)}{\partial x^m}
\end{bmatrix}^\top$, we can create an equivalent state space form with the process model as
\begin{equation}
    \begin{aligned}\label{eqn:SSGP_def}
 \frac{\partial  \mathbf{f}(x)}{\partial x}&=\underbrace{\begin{bmatrix}
0 & 1 & & \\
& \ddots & \ddots & \\
& & 0 & 1 \\
-a_{0} & -a_{1} & \cdots & -a_{m-1}
\end{bmatrix}}_{\mathbf{A}} \mathbf{f}(x) +\underbrace{\begin{bmatrix}
0 \\
\vdots \\
0 \\
1
\end{bmatrix}}_{\mathbf{L}} Q(x),\\
y&=\underbrace{\begin{bmatrix}
1 & 0 & \cdots & 0
\end{bmatrix}}_{\mathbf{H}} \mathbf{f}(x)+\varepsilon,
\end{aligned}
\end{equation}
For the output equation recall that eq.~(\ref{eq_GP_measurements}) is already in linear form therefore no modification is needed.
Once this state space equation is derived, a Kalman Filter or other similar technique can be employed to linearly iterate across the dataset as a sequence to make a prediction. To further elaborate, given a random input $x_\ast$, and an initial state $\mathbf{f}(x)$, the system will iteratively move through the dataset $\mathcal{D}$ performing Kalman predictions and updates till it updates on $x_i \in \mathcal{D}$ which is the closest data point to $x_\ast$. Consequently, the derivatives such as $\frac{\partial  \mathbf{f}(x)}{\partial x}$ and higher are updated during this motion allowing the system to observe a data trend. This facilitates a final prediction between $x_i$ and $x_\ast$ based on the previously observed trend. Consequently, our system is no longer deterministic  as the system can either approach from the right or left of $x_\ast$.
Extending the formulation to multiple input single output (MISO), can be obtained by stacking the eq.~(\ref{eqn:SSGP_def}) diagonally for the $\mathbf{A}$ matrix, horizontally $\mathbf{H}$, and vertically for $\mathbf{L}$
\begin{equation}
        \begin{aligned}
\ \begin{bmatrix}\frac{\partial  \mathbf{f}_0(x_0)}{\partial  t}\\ \vdots\\
\frac{\partial  \mathbf{f}_d(x_d)}{\partial  t}\end{bmatrix}&=\begin{bmatrix} \mathbf{A}_{0}& ...& \mathbf{0}\\   & \ddots&\\
\mathbf{0} & ... & \mathbf{A}_{d}\end{bmatrix}\begin{bmatrix}\mathbf{f}_0(x_0)\\ \vdots\\
\mathbf {f}_d(x_d)\end{bmatrix}+\begin{bmatrix} \mathbf{L}_{0}\\ \vdots \\ \mathbf{L}_{d} \end{bmatrix}Q(\mathbf{x}),\\
y&=\begin{bmatrix}\mathbf{H}_{0}  & ... &\mathbf{H}_{d}\end{bmatrix}\begin{bmatrix}\mathbf{f}_0(x_0)\\ \vdots\\
\mathbf {f}_d(x_d)\end{bmatrix}+\epsilon. \end{aligned}
\end{equation}
Consequently, the conversion of kernel functions into linear state space forms expressed as matrices allows easy exportation to other frameworks. This is because there is no need to write additional functions simply swapping transition matrix parameters is sufficient, and consequently dropping any library dependencies needed to solve for the transition matrices. Our toolbox has this functionality which allows exporting python models directly to C++.

\subsection{Kernel Formulation}
 
We give example derivations for our toolbox's kernel using Algorithm \ref{alg_ssgp}. 
\subsubsection{Radial Basis Function Kernel}
The Radial Basis Function (RBF) kernel is also known as the squared exponential kernel. No closed-form solution for any arbitrary approximation order $m$ exists. Fortunately, following algorithm \ref{alg_ssgp}, an appropriate approximation can always be derived for some order $m$ and hyperparameter $z \in \mathbb{R}_{>0}$ known as the lengthscale. For future reference $\mathbb{R}_{>0}$ is the set of all positive real numbers excluding zero
\begin{equation}
        \begin{split}
    k(\Delta) &=e^{-0.5z^2\Delta^2},\\
    s(j\omega)s(-j\omega) &= 1+(z\omega)^2+...\frac{1}{m!}(z\omega)^{2m}.\\
 \end{split}
\end{equation}
Considering $m=2$ and $z=1$ we obtain
\begin{equation}
        \begin{split}
     &s(j\omega)s(-j\omega) = 1+(\omega)^2+0.5(\omega)^{4},\\
   &= 0.5(\omega^2+(1+j))(\omega^2+(1-j)),\\
       s(j\omega)&= \frac{1}{\sqrt{2}}(j\omega+\sqrt{1+j})(j\omega+\sqrt{1-j}),\\
    s(j\omega)&\approx \frac{1}{\sqrt{2}}(j\omega)^2+\frac{2.2}{\sqrt{2}}(j\omega)+1.
 \end{split}
\end{equation}

\subsubsection{Matern Kernel}
Matern is a more complex and generalizable kernel compared to the RBF Kernel and approaches the RBF Kernel as $\nu\rightarrow\infty$. This kernel is one of the few kernels where there is a closed-form solution for any arbitrary set of hyperparameters 

\begin{equation}
        \begin{aligned}
k(\Delta)&=\sigma^2 \frac{2^{1-\nu}}{\Gamma(\nu)}\left(\sqrt{2 \nu} \frac{\Delta}{z}\right)^\nu K_\nu\left(\sqrt{2 \nu} \frac{\Delta}{z}\right),\\    s(j\omega)s(-j\omega) &=  \left(\lambda^2+\omega^2\right)^{(\nu+1 / 2)},\\
s(j\omega) &=(\lambda+j \omega)^{(\nu+1 / 2)},
 \end{aligned}
\end{equation}
where $\lambda = \frac{\sqrt{2\nu}}{z}$. $\sigma,\nu,z \in \mathbb{R}_{>0}$ are hyperparameters of the kernel. $\Gamma$ and $K_\nu$ are the gamma function and modified Bessel function of the second kind respectively. 

\subsubsection{Periodic Kernel}The periodic kernel has good expressiveness with oscillating features
\begin{equation}
        \begin{split}
    k(\Delta) &=e^{-2z^2sin^2(0.5w_0\Delta^2)},\\
    s(j\omega)s(-j\omega) &= \sum_{j=0}^\infty q_j^2(\delta(\omega-j\omega_0)+\delta(\omega+j\omega_0)).
 \end{split}
\end{equation}
Unfortunately, the power spectrum is a summation of impulse functions, $\delta$. It is possible to obtain a slightly modified state space approximating the series to the limit $J \rightarrow \infty$ \cite{solin2014explicit}
\begin{equation}
A_j^k = \begin{bmatrix} 0 & -j\omega_{0}\\   
j\omega_{0} & 0\end{bmatrix}.
\end{equation}
The measurement matrix $\mathbf{H}$ block row vector is a modified $\mathbf{H}_j^k = [1~0]$, while the noise diffusion is the standard $\mathbf{L}_j^k =[0~1]^\top$. To acquire a better approximation, the system $_k$ matrices are stacked similarly to the MISO system $k$ times. 

%% file: Sections/timing_table_average.tex
\begin{table*}[t]
    \centering
    \caption{\label{tab:laptop_evaluation}}
    \vspace{-0.75em}
    \caption*{\scshape Average Inference Time per Data point [$\SI{}{\milli\second}$]}
    \vspace{-0.3em}
    \begin{tabular}{c r r r r r r r r r r r r}
        \toprule\toprule
        \multirow{3}{*}{Points} &
        \multicolumn{3}{c}{SISO SSGP} &
        \multicolumn{3}{c}{SISO GPT} &
        \multicolumn{3}{c}{MISO SSGP} &
        \multicolumn{3}{c}{MISO GPT}\\
        \cmidrule(lr){2-4}\cmidrule(lr){5-7}\cmidrule(lr){8-10}\cmidrule(lr){11-13}
        & \multicolumn{1}{c}{PER} & \multicolumn{1}{c}{RBF} & \multicolumn{1}{c}{MAT}
        & \multicolumn{1}{c}{PER} & \multicolumn{1}{c}{RBF} & \multicolumn{1}{c}{MAT}
        & \multicolumn{1}{c}{PER} & \multicolumn{1}{c}{RBF} & \multicolumn{1}{c}{MAT}
        & \multicolumn{1}{c}{PER} & \multicolumn{1}{c}{RBF} & \multicolumn{1}{c}{MAT}\\
        \midrule
$10$   & $5.77$ & $3.57$ & $3.36$ & $2.01$ & $0.96$ & $0.99$ &      $8.11$  & $6.09$ & $5.79$   & $1.68$ & $0.97$ & $1.01$ \\
$50$   & $5.74$ & $3.51$ & $3.34$ & $1.73$ & $1.01$ & $1.10$ &     $30.18$ & $27.85$ & $25.96$ & $1.71$ & $0.97$ & $1.06$ \\
$200$  & $5.83$ & $3.53$ & $3.37$ & $2.09$ & $1.18$ & $1.33$ &     $32.35$ & $30.40$ & $28.29$ & $2.29$ & $1.21$ & $1.33$ \\
$1000$ & $6.61$ & $4.43$ & $4.27$ & $16.04$ & $3.23$ & $4.16$ &    $43.23$ & $41.21$ & $39.44$ & $46.45$ & $3.16$ & $4.95$ \\
$2500$ & $5.71$ & $3.61$ & $3.48$ & $93.64$ & $34.24$ & $57.71$ &   $57.69$ & $55.76$ & $53.77$ & $237.86$ & $33.98$ & $57.32$ \\
$6000$ & $6.91$ & $4.50$ & $4.43$ & $402.47$ & $167.29$ & $255.39$ & $88.19$ & $86.26$ & $84.26$ & $1040.41$ & $154.40$ & $254.38$ \\
        \bottomrule\bottomrule
    \end{tabular}
    \vspace{-2em}
    \label{tab:inference_time}
\end{table*}

%% file: Sections/V_experimental_setup.tex
\section{Learning Quadrotor Dynamics}\label{sec:quad_dynamics}
We validate our approach by considering a learning residual quadrotor dynamics problem. In the following section, we describe the formulation of our problem for both a Single input single output (SISO) GP implementation and MISO GP formulation. For our SISO GP, we follow a similar modeling approach to \cite{Guillem_datadriveMPC}. We consider the following dynamics
\begin{equation}\label{eq:nominal_dynamic}
m\mathbf{a} = \mathbf{R}^W_B\tau \mathbf{e}_3 - mg\mathbf{e}_3, 
\end{equation}
where $\mathbf{e}_3 = [0~0~1]^\top$, $\mathbf{a}$ is the acceleration in the world frame of the quadrotor, $g$ the gravitational value, and $\mathbf{R}^W_B$ the rotation converting points from the quadrotor's body frame to the world frame. The vehicle's velocity in the body frame, $\mathbf{v}_b$ is used to estimate the acceleration error on the body of the vehicle $\bm{\delta}_a$. This is based on standard fluid dynamic approximations of the air resistance for a moving body when evolving at different speeds. In our dataset, we calculate the term of $\bm{\delta}_a$ as
\begin{equation}
     \bm{\delta}_a=\mathbf{R}^W_B\left(\frac{\mathbf{v}_b - \hat{\mathbf{v}}_b}{\delta t_k}\right),
\end{equation}
where $\hat{\mathbf{v}}_b$ is the predicted velocity from integrating eq.~(\ref{eq:nominal_dynamic}) from the previous time step, and $\delta t_k$ is the length of that time step. We can then construct our SISO GP using $\mathbf{v}_b$ as the input and $\bm{\delta}_a$ as the output, acceleration error, we would like to predict. To handle multiple outputs, we consider each body axis independent and apply an independent GP individually to each axis component. This makes our augmented GP dynamic formulation of eq.~(\ref{eq:nominal_dynamic}) as
\begin{equation}\label{eq:siso_drag}
        \mathbf{a} = \frac{1}{m}\mathbf{R}^W_B(\tau \mathbf{e_3} +f_{GP}(\mathbf{v}_b))+g\mathbf{e_3}.
\end{equation}
Fundamentally, the GP term $f_{GP}$ in eq.~(\ref{eq:siso_drag}) infers a first order propeller drag term~\cite{Guillem_datadriveMPC,svacha}. However, the vehicle is generally subject to additional types of disturbances. These disturbances are a function of motor thrust which is a function of $\omega_i$ of the motor's angular velocity through a $k_f$ coefficient, obtaining the total thrust as
\begin{equation}\label{eqn:motor_dynamics}
        \begin{aligned}
    \tau &=T_0 + T_1 + T_2 + T_3,~T_{i} = \omega_i^2*k_f,
   \end{aligned}
\end{equation}
where $T_{i}$ is the motor thrust generated by the motor $i$.  
However, the thrust mapping is only an approximation and does not account for other external factors or changes in the propeller over time~\cite{DAI_propeller_analysis}. Our system augments this term by using a MISO GP to calculate the disturbances related to motor thrust from both the fluid resistance from high body rate changes and inaccuracies in the thrust mapping. Therefore, our full system augmented model of eq.~(\ref{eq:nominal_dynamic}) for MISO becomes
\begin{equation}
        \mathbf{a} = \frac{1}{m}\mathbf{R}^W_B(\tau \mathbf{e_3} +f_{GP}(\bm{v}_b, \omega_0,\omega_1,\omega_2,\omega_3))+g\mathbf{e_3}.
\end{equation}

%% file: Sections/pred_performance_fig.tex
\begin{figure*}
    \centering
\includegraphics[width=\linewidth, trim=0 0 0 10cm, clip]{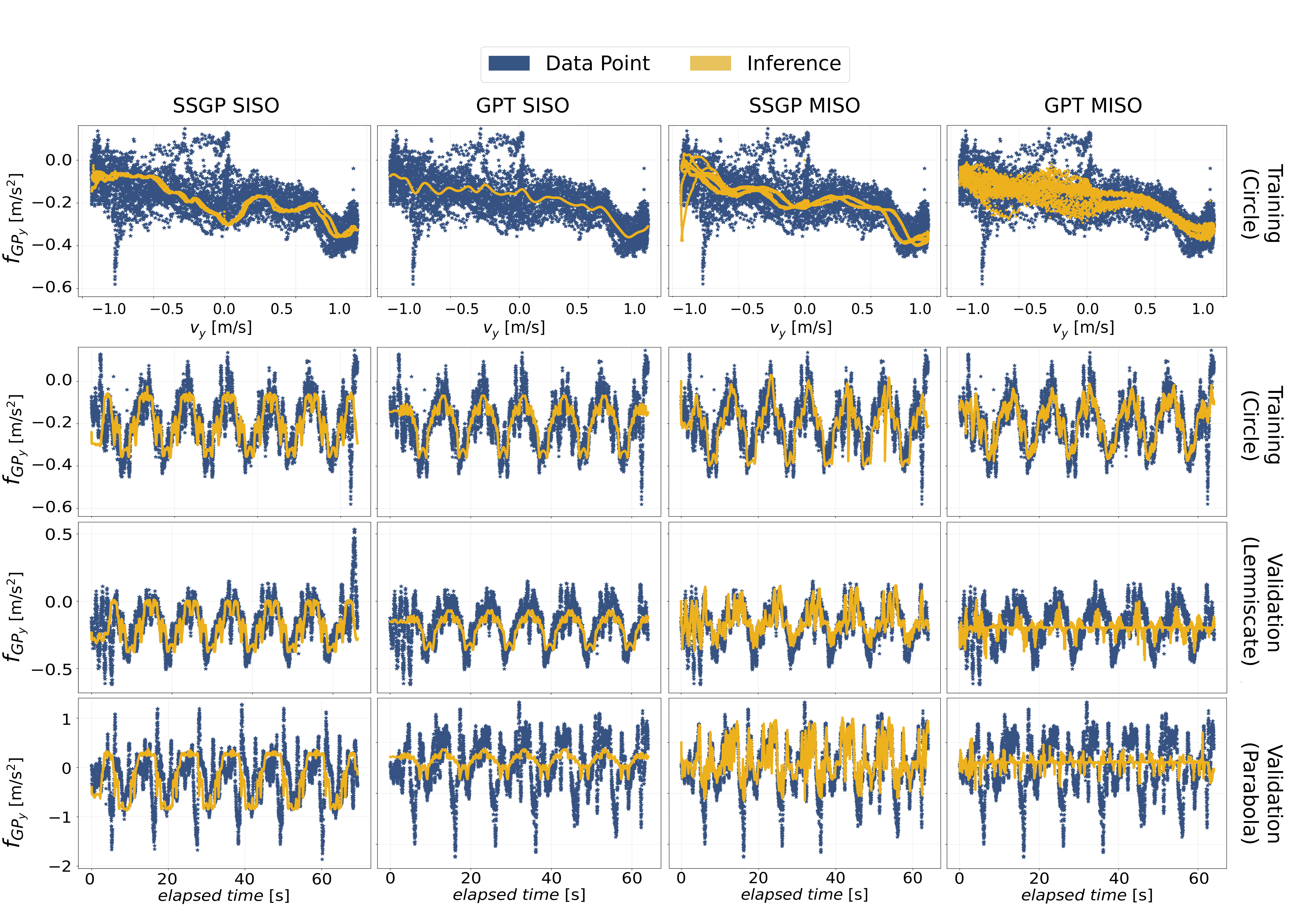}
\caption{Predictive performance of the propeller drag term on the y-axis of $f_{GP}$ denoted as $f_{{GP}_y}$ over training and validation datasets.  Data trend between body velocity and residual acceleration error (first row). Visualization of the fitting on the residuals with respect to time (second to fourth rows).}
    \label{fig:pred_performance}
    \vspace{-10pt}
\end{figure*}

%% file: Sections/VI_results.tex
\section{Results}
We design our evaluation procedure to address the following questions. 
(i) What is the computational speed gain from converting to state space versus a traditional matrix inversion eq.~(\ref{eq_GP_prediction})? 
(ii) Can the SSGP perform robust inference to the real-world data comparable to the baseline matrix inversion implementation of eq.~(\ref{eq_GP_prediction})? 
(iii) Can extending the input space to multiple inputs greatly improve the inference?
We perform several experiments to address these questions. First, we compare our computational time for solving a full dataset between our implementation and a traditional method. Next, we compared the inference RMSE and fitting of our SSGP approach with a traditional inference approach of matrix inversion for both SISO and MISO. 
For our baseline implementation of eq.~(\ref{eq_GP_prediction}) we leverage GPytorch (GPT), \cite{gardner2018gpytorch} a library that performs robust and efficient matrix operations and optimizations by integrating with Pytorch's ecosystem. 
This allows GPT to perform both efficient optimization of hyperparameters for the objective eq.~(\ref{eq_log_marg_likelihood}) and inference with eq.~(\ref{eq_GP_prediction}) . We evaluated our approach on a real-world dataset of quadrotor dynamics and flight, \cite{Ale_PiTCN}. This dataset contains real-world unfiltered data resulting in much noisier signals and stressing our models' ability to perform inference. We chose three different flight trajectories: Circle (training), Parabola (validation), and Lemniscate (validation).
To validate inference time, we tested our implementation on a 11th Gen Intel i7 for both  SSGP and GPT in Table~\ref{tab:inference_time} with pure CPU computation. The results were repeated with a Xavier NX achieving similar timing trends. This table records the average amount of time taken to complete one inferences given a dataset of size $n$. Our time comparison consists of the following kernels: RBF, Matern, and periodic given different approximation orders, RBF $m=6$, Matern $m=3$, and periodic $m=18$ and compared with both GPT in a SISO format and a MISO ($5$ independent inputs). Our model for SISO or $m\leq10$ can perform inference much faster than a comparable baseline implementation, especially for datasets $>1000$. However, the SSGP formulation slows down in the MISO formulation. This is because computationally MISO for the SSGP of order $m=6$ and $5$ inputs is equivalent to a $m=30$ approximation. This gives even slower computations than the GPT version until  $2500+$ points are used. 
\begin{table}[t]
    \centering
    \caption{\label{tab:laptop_evaluation}}
    \vspace{-0.75em}
    \caption*{\scshape Predictive Performance RMSE [$\SI{}{\meter\per\second\squared}$]}
    \vspace{-0.3em}
    \begin{tabular}{r r r r r}
        \toprule\toprule
        \multirow{2}{*}{Trajectory} &
        \multicolumn{2}{c}{SISO} &
        \multicolumn{2}{c}{MISO} \\   
        \cmidrule(lr){2-3}\cmidrule(lr){4-5}
        & \multicolumn{1}{c}{SSGP} & \multicolumn{1}{c}{GPT} & \multicolumn{1}{c}{SSGP} &  \multicolumn{1}{c}{GPT}  \\
        \midrule
        Circle (training) & $0.0938$ & $0.0755$& $0.0852$&  $\mathbf{0.0666}$\\
        Transposed Parabola (val) & $0.1330$& $0.1164$ &$\mathbf{0.0889}$  &$0.1516$\\
        Extended Lemniscate (val) & $0.6079$& $0.4561$& $\mathbf{0.3176}$ & $0.4698$\\
        \bottomrule\bottomrule
    \end{tabular}
    \label{tab:rmse_performances}
\vspace{-20pt}
\end{table}
Our inference task on the dataset is to model the residual accelerations defined in Section \ref{sec:quad_dynamics} with body velocity as the input to prediction acceleration error for SISO, and body velocity and motor speeds for MISO.
For our RMSE evaluation in Fig.~(\ref{fig:pred_performance}), we focused on estimating the drag term on the y-axis, $f_{{GP}_y}$, choosing the RBF Kernel with an approximation order $m=6$.  The residual inference performance of SSGP is compared to the same kernel implemented in GPT using the same hyperparameters. As observed, our SSGP method despite performing similarly to GPT on the training set better generalizes on the testing dataset with lower RMSEs across all test sets (see Table \ref{tab:rmse_performances}), and can perform faster inference. We believe the greater generalizability of our system depends on the prior induced by the SSGP structure where inference is achieved through observing the data trend rather than a correlation on the dataset like in GPT. Furthermore scaling our inputs to include the motor rates allows the MISO SSGP to better capture high-frequency features than a SISO system.
Similar results are obtained for the other drag components $f_{{GP}_x}$, $f_{{GP}_z}$.

%% file: Sections/VII_conclusions.tex
\section{Conclusion}
In this work, we presented  GaPT, a scalable and fast toolkit for state space GPs. This method shows superior computational performance compared to a classical GP inference approach on both single and multi-input settings especially for large datasets. \projectName~ enables easy exporting of GPs into C++ functions allowing direct integration in common robotics libraries enabling real-time inference speed even on embedded platforms used on SWaP-constrained aerial robots. \projectName~ incorporates both single of multiple inputs solutions. The latter further improves the inference capabilities of the proposed approach. Future works will involve adding additional kernels into the Toolkit such as spectral mixture or deep state-space models which can better capture complex patterns. This process will also be facilitated by the open source implementation that the robotics community can leverage to collaborate by adding additional functionalities that may be helpful for different purposes. Furthermore, we will show the integration of our method in various additional robotic use cases such as for control in aggressive maneuvers~\cite{mao2022robust}, calculating safety sets, or exploiting the full power of model predictive control with online refined learned dynamics.

%% file: main.bbl
\begin{thebibliography}{10}
\providecommand{\url}[1]{#1}
\csname url@rmstyle\endcsname
\providecommand{\newblock}{\relax}
\providecommand{\bibinfo}[2]{#2}
\providecommand\BIBentrySTDinterwordspacing{\spaceskip=0pt\relax}
\providecommand\BIBentryALTinterwordstretchfactor{4}
\providecommand\BIBentryALTinterwordspacing{\spaceskip=\fontdimen2\font plus
\BIBentryALTinterwordstretchfactor\fontdimen3\font minus
  \fontdimen4\font\relax}
\providecommand\BIBforeignlanguage[2]{{%
\expandafter\ifx\csname l@#1\endcsname\relax
\typeout{** WARNING: IEEEtran.bst: No hyphenation pattern has been}%
\typeout{** loaded for the language `#1'. Using the pattern for}%
\typeout{** the default language instead.}%
\else
\language=\csname l@#1\endcsname
\fi
#2}}

\bibitem{Guillem_datadriveMPC}
G.~Torrente, E.~Kaufmann, P.~Föhn, and D.~Scaramuzza, ``Data-driven mpc for
  quadrotors,'' \emph{IEEE Robotics and Automation Letters}, vol.~6, no.~2, pp.
  3769--3776, 2021.

\bibitem{hewing2019cautious}
L.~Hewing, J.~Kabzan, and M.~N. Zeilinger, ``Cautious model predictive control
  using gaussian process regression,'' \emph{IEEE Transactions on Control
  Systems Technology}, vol.~28, no.~6, pp. 2736--2743, 2019.

\bibitem{Deisenroth_GP_robotics}
M.~P. Deisenroth, D.~Fox, and C.~E. Rasmussen, ``Gaussian processes for
  data-efficient learning in robotics and control,'' \emph{IEEE Transactions on
  Pattern Analysis and Machine Intelligence}, vol.~37, no.~2, pp. 408--423,
  2015.

\bibitem{barfoot2014batch}
T.~D. Barfoot, C.~H. Tong, and S.~S{\"a}rkk{\"a}, ``Batch continuous-time
  trajectory estimation as exactly sparse gaussian process regression.'' in
  \emph{Robotics: Science and Systems}, vol.~10.\hskip 1em plus 0.5em minus
  0.4em\relax Citeseer, 2014, pp. 1--10.

\bibitem{Anayo_reachabilityonlineGP}
A.~K. Akametalu, J.~F. Fisac, J.~H. Gillula, S.~Kaynama, M.~N. Zeilinger, and
  C.~J. Tomlin, ``Reachability-based safe learning with gaussian processes,''
  in \emph{53rd IEEE Conference on Decision and Control}, 2014, pp. 1424--1431.

\bibitem{WangonlineGP2018}
L.~Wang, E.~A. Theodorou, and M.~Egerstedt, ``Safe learning of quadrotor
  dynamics using barrier certificates,'' in \emph{IEEE International Conference
  on Robotics and Automation (ICRA)}, 2018, pp. 2460--2465.

\bibitem{loquercio2022}
A.~Loquercio, A.~Saviolo, and D.~Scaramuzza, ``Autotune: Controller tuning for
  high-speed flight,'' \emph{IEEE Robotics and Automation Letters}, vol.~7,
  no.~2, pp. 4432--4439, 2022.

\bibitem{gardner2018gpytorch}
J.~Gardner, G.~Pleiss, K.~Q. Weinberger, D.~Bindel, and A.~G. Wilson,
  ``Gpytorch: Blackbox matrix-matrix gaussian process inference with gpu
  acceleration,'' \emph{Advances in neural information processing systems},
  vol.~31, 2018.

\bibitem{acados}
R.~Verschueren, G.~Frison, D.~Kouzoupis, N.~van Duijkeren, A.~Zanelli,
  R.~Quirynen, and M.~Diehl, ``Towards a modular software package for embedded
  optimization,'' in \emph{Proceedings of the IFAC Conference on Nonlinear
  Model Predictive Control (NMPC)}, 2018.

\bibitem{FORCESPro}
\BIBentryALTinterwordspacing
A.~Domahidi and J.~Jerez. (2014--2019) Forces professional. Embotech AG.
  [Online]. Available: \url{https://embotech.com/FORCES-Pro}
\BIBentrySTDinterwordspacing

\bibitem{mastalli20crocoddyl}
C.~Mastalli, R.~Budhiraja, W.~Merkt, G.~Saurel, B.~Hammoud, M.~Naveau,
  J.~Carpentier, L.~Righetti, S.~Vijayakumar, and N.~Mansard, ``{Crocoddyl: An
  Efficient and Versatile Framework for Multi-Contact Optimal Control},'' in
  \emph{IEEE International Conference on Robotics and Automation (ICRA)}, 2020.

\bibitem{biyik2020active}
E.~B{\i}y{\i}k, N.~Huynh, M.~J. Kochenderfer, and D.~Sadigh, ``Active
  preference-based gaussian process regression for reward learning,'' in
  \emph{Robotics: Science and System XVI}, 2020.

\bibitem{gahlawat2020l1}
A.~Gahlawat, P.~Zhao, A.~Patterson, N.~Hovakimyan, and E.~Theodorou, ``L1-gp:
  L1 adaptive control with bayesian learning,'' in \emph{Learning for Dynamics
  and Control}.\hskip 1em plus 0.5em minus 0.4em\relax PMLR, 2020, pp.
  826--837.

\bibitem{matthews2017gpflow}
A.~G. d.~G. Matthews, M.~Van Der~Wilk, T.~Nickson, K.~Fujii, A.~Boukouvalas,
  P.~Le{\'o}n-Villagr{\'a}, Z.~Ghahramani, and J.~Hensman, ``Gpflow: A gaussian
  process library using tensorflow.'' \emph{J. Mach. Learn. Res.}, vol.~18,
  no.~40, pp. 1--6, 2017.

\bibitem{SMkernerl}
G.~Parra and F.~Tobar, ``Spectral mixture kernels for multi-output gaussian
  processes,'' in \emph{Proceedings of the 31st International Conference on
  Neural Information Processing Systems}, ser. NIPS'17.\hskip 1em plus 0.5em
  minus 0.4em\relax Red Hook, NY, USA: Curran Associates Inc., 2017, p.
  6684–6693.

\bibitem{zhao2021deep}
Z.~Zhao, M.~Emzir, and S.~S{\"a}rkk{\"a}, ``Deep state-space gaussian
  processes,'' \emph{Statistics and Computing}, vol.~31, no.~6, pp. 1--26,
  2021.

\bibitem{Sivaram_fastGP_calc}
S.~Ambikasaran, D.~Foreman-Mackey, L.~Greengard, D.~W. Hogg, and M.~O’Neil,
  ``Fast direct methods for gaussian processes,'' \emph{IEEE Transactions on
  Pattern Analysis and Machine Intelligence}, vol.~38, no.~2, pp. 252--265,
  2016.

\bibitem{joaquin_unifiyng_view}
J.~Qui{{\~n}}onero-Candela and C.~E. Rasmussen, ``A unifying view of sparse
  approximate gaussian process regression,'' \emph{Journal of Machine Learning
  Research}, vol.~6, no.~65, pp. 1939--1959, 2005.

\bibitem{DAS201812}
``Fast gaussian process regression for big data,'' \emph{Big Data Research},
  vol.~14, pp. 12--26, 2018.

\bibitem{sarkka2011linear}
S.~S{\"a}rkk{\"a}, ``Linear operators and stochastic partial differential
  equations in gaussian process regression,'' in \emph{International Conference
  on Artificial Neural Networks}.\hskip 1em plus 0.5em minus 0.4em\relax
  Springer, 2011, pp. 151--158.

\bibitem{corenflos2022temporal}
A.~Corenflos, Z.~Zhao, and S.~S{\"a}rkk{\"a}, ``Temporal gaussian process
  regression in logarithmic time,'' in \emph{25th International Conference on
  Information Fusion (FUSION)}, 2022, pp. 1--5.

\bibitem{sarkka2013spatiotemporal}
S.~Sarkka, A.~Solin, and J.~Hartikainen, ``Spatiotemporal learning via
  infinite-dimensional bayesian filtering and smoothing: A look at gaussian
  process regression through kalman filtering,'' \emph{IEEE Signal Processing
  Magazine}, vol.~30, no.~4, pp. 51--61, 2013.

\bibitem{RasmussenW06}
C.~E. Rasmussen and C.~K.~I. Williams, \emph{Gaussian processes for machine
  learning.}, ser. Adaptive computation and machine learning.\hskip 1em plus
  0.5em minus 0.4em\relax MIT Press, 2006.

\bibitem{murray2010slice}
I.~Murray and R.~P. Adams, ``Slice sampling covariance hyperparameters of
  latent gaussian models,'' \emph{Advances in neural information processing
  systems}, vol.~23, 2010.

\bibitem{hartikainen2010kalman}
J.~Hartikainen and S.~S{\"a}rkk{\"a}, ``Kalman filtering and smoothing
  solutions to temporal gaussian process regression models,'' in \emph{IEEE
  international workshop on machine learning for signal processing}, 2010, pp.
  379--384.

\bibitem{bochner1959lectures}
S.~Bochner \emph{et~al.}, \emph{Lectures on Fourier integrals}.\hskip 1em plus
  0.5em minus 0.4em\relax Princeton University Press, 1959, vol.~42.

\bibitem{stein1999interpolation}
M.~L. Stein, \emph{Interpolation of spatial data: some theory for
  kriging}.\hskip 1em plus 0.5em minus 0.4em\relax Springer Science \& Business
  Media, 1999.

\bibitem{chatfield2003analysis}
C.~Chatfield, \emph{The analysis of time series: an introduction},
  6th~ed.\hskip 1em plus 0.5em minus 0.4em\relax CRC Press, 2004.

\bibitem{solin2014explicit}
A.~Solin and S.~S{\"a}rkk{\"a}, ``Explicit link between periodic covariance
  functions and state space models,'' in \emph{Artificial Intelligence and
  Statistics}.\hskip 1em plus 0.5em minus 0.4em\relax PMLR, 2014, pp. 904--912.

\bibitem{svacha}
J.~Svacha, J.~Paulos, G.~Loianno, and V.~Kumar, ``Imu-based inertia estimation
  for a quadrotor using newton-euler dynamics,'' \emph{IEEE Robotics and
  Automation Letters}, vol.~5, no.~3, pp. 3861--3867, 2020.

\bibitem{DAI_propeller_analysis}
X.~Dai, Q.~Quan, J.~Ren, and K.-Y. Cai, ``An analytical design-optimization
  method for electric propulsion systems of multicopter uavs with desired
  hovering endurance,'' \emph{IEEE/ASME Transactions on Mechatronics}, vol.~24,
  no.~1, pp. 228--239, 2019.

\bibitem{Ale_PiTCN}
A.~Saviolo, G.~Li, and G.~Loianno, ``Physics-inspired temporal learning of
  quadrotor dynamics for accurate model predictive trajectory tracking,''
  \emph{IEEE Robotics and Automation Letters}, vol.~7, no.~4, pp.
  10\,256--10\,263, 2022.

\bibitem{mao2022robust}
J.~Mao, S.~Nogar, C.~M. Kroninger, and G.~Loianno, ``Robust active visual
  perching with quadrotors on inclined surfaces,'' \emph{IEEE Transactions on
  Robotics}, pp. 1--17, 2023.

\end{thebibliography}
